\documentclass[11pt,a4paper]{article}
\usepackage[hyperref]{acl2017}
\usepackage{times}
\usepackage{latexsym,comment}
\usepackage{url}
\usepackage{fancyhdr}

\aclfinalcopy

\title{Towards the Study of Morphological Processing of the Tangkhul Language}
\author{Mirinso Shadang \qquad Navanath Saharia \qquad Thoudam Doren Singh \\ Indian Institute of Information Technology Manipur\\
Imphal, India - 795002\\
{\{\tt mirinso, nsaharia\}@iiitmanipur.ac.in,thoudam.doren@gmail.com} 
}

\date{}
\begin{document}		
\maketitle

\begin{abstract}
There is no or little work on natural language processing of Tangkhul language. The current work is a humble beginning of morphological processing of this language using an unsupervised approach. We use a small corpus collected from different sources of text books, short stories and articles of other topics. Based on the experiments carried out, the morpheme identification task using morphessor gives reasonable and interesting output despite using a small corpus.
                
\end{abstract}                                          

\section{Introduction}
The name Tangkhul (also known as Hao or Ihao especially in older literature) refers to an ethnic group which live in the hill of Ukhrul district  and Kamjong district of Manipur State, India, and contiguous parts of Nagaland (another state of India) and Burma. The Tangkhuls are quite diversified linguistically, and culturally. The speech varieties of most Tangkhul villages are not mutually intelligible with those of neighbouring villages. Even the similarities are large enough to facilitate the rapid learning of one another’s languages. However, it is clear that the Tangkhul languages are closely related to one-another and form a distinct subgroup within the Tibeto-Burman family. Tangkhul is an ethnic group whose language varies from village to village.  It has long been noted that Tangkhul is a group of languages, rather than a single language (Brown 1837), however, almost all of the available descriptions of Tangkhul languages have concentrated on a single variety—the language of Ukhrul town, which has come to serve as a lingua franca for the whole Tangkhul tribe. Even if the Tangkhul language serve as a common language for the whole Tangkhul region, but the accent of speech (way of speaking Tangkhul) varies from geographical area. Tangkhul region is mainly divided into four major part i.e. East (zingsho), west (zingtun), North (ato), South (azay). There are more than 100 language spoken in the various villages of Tangkhul diaspora, since all the villages of Tangkhul region (Ukhrul district \& Kamjong district) has their own native language, the pronunciation of the common language (Tangkhul) is highly varying.

The most obvious working definition of the Tangkhul language is the family of Tibeto-Burman languages spoken by members of the Tangkhul tribe. This definition is complicated by a number of factors, and is clearly inadequate (as will be seen), since some of the languages spoken as a mother-tongue by ethnic Tangkhuls are not members of the family being discussed here  and because it is possible that there are members of other Naga tribes speaking languages that belong in the Tangkhul group. Thus, while ethnicity can be taken as neither a necessary nor a sufficient criterion for membership in the Tangkhul family it is nevertheless a useful starting point for a discussion of this group of languages.

Morphological processing of a language is the beginning of NLP work of that particular language. There are reports on morphological processing work for many other major Indian languages. But, Tangkhul language in particular which falls under resource constrained and less privileged category with no or little work on natural language processing till date. In this light, we attempt to collect small corpus and start the initial steps towards morphological processing of this language. The morphological analysis of this language is found to be slightly complex being agglutinative~\cite{mortensen2003comparative} one.

\section{Related Work}

The Tibeto-Burman languages of India are still lacking the basic language processing tools of required quality. Among the reported work on 
Tibeto-Burman languages, Bodo~\cite{sarmah2004}, Mizo~\cite{pakray15}, Kok-borok~\cite{debbarma2012} and Manipuri~\cite{singh2008} are main. Almost all the reported work used apriori knowledge of the language either in the form of dictionary or in the form of preparing label data for training. Being a resource less language, we preferred unsupervised learning approach for the considered language.  

\subsection{Language morphology and syntax}

Tankhul has two distinct tones - high and low, mainly with the utterances associated with the letter t, d, r, m, and f.  
%
For examples
\begin{enumerate}
\item High (Kajuiya)

                           Chanhan                        Kasho (open)
                            Khaikao (dry fish)            Khalen (trap)

\item Low (Khanema)
                      kachang (month)               kachui (high)
                      mashit (closeness)             ashee (blood)
\end{enumerate}

Adjective of Tangkhul displays a hybrid morphology, as it always occurs semantically somewhere between verbs and nouns.

Like other morphologically rich neighbour languages~\cite{majumder2007,saharia2014}, Tangkhul also has a strong tendency to form sequence of affixes to prepend/append to the root.  


In fact, many of the prefixes found in Tangkhul language behaves morphologically as if they are part of the root. While it is possible to assign independent historical origins to these elements, from a synchronic standpoint the factors that identify them are phonological and not morphological. As will be seen, it is often not possible to assign a consistent meaning or grammatical function to these prefixes, which will be referred to here as lexical prefixes~~\cite{mortensen2003comparative}. 

\subsection{Word Formation}
The formation of words in a language is one the topic, which still attracts the researchers most. The two main class of word formation are inflection and derivation. Derivation is divided into again two classes - affixation and compounding. Compounding is divided into three classes - endocentric compound, exocentric compound and co-ordinate compound. Endocentric compound is again divided into right headed compound and left-headed compound. In the right-headed compound, it may have many constituents such as `noun + noun' and `noun + derived noun', but `noun + noun' is more used than the other one in Tangkhul \cite{ahum1997tangkhul}. 


There are three types of verb roots in Tangkhul- simple, complex and compound.
%
A simple root is irreducible 'core element' obtained by dropping all the affixes. A root may be monosyllabic or bisyllabic. For example: lei ({\em exist/have}),  vao ({\em shout}),  and malai ({\em forget}). 
%
Though Tangkhul verbs are close class,  
few verbs are derived from 'nominal' roots. For example: /pha/ /ra/. Table~\ref{caseMarker} tabulated few case markers of Tanghul language.

\begin{table*}[!htbp]
\centering
\begin{tabular}{@{}lll@{}} 
\hline
Case marker&			Suffix                  &   Example\\
Locative case marker		& -li, -wui		&	         Manipur-li (in Manipur)\\
Genitive case marker	  &              -chi	-        &                            Avi-chi (his)\\
Nominative case marker    &             -na &                                           khipa-na (who)\\
\hline
\end{tabular}
\caption{\label{caseMarker} Few case markers with example}
\end{table*}
…


There are two broad types of modifiers- adjectival and adverbials. 
%
Adjectives, as a word class, are quite different from nouns and verbs. In the case of Tangkhul language, the exact relationship between adjectives on the one hand, and other categories like nouns, verbs and adverbs on the other, has been one of the disputed issues in linguistics. In Tangkhul language there is no distinction between verbs and adjectives in the sense that they are derived from roots, and function as adjectives or verbs with (a) appropriate affixation and (b) appropriate occurrence in a sentence~\cite{ahum1997tangkhul}. 
%
Expressives (which are aplenty in the language and which most often have adverbial and adjectival functions) can be compounded with a number of roots to form compound adjectival. In the process of compounding expressive tend to behave like intensifiers or modifiers. The following are some of the most commonly used adjectives formed by compounding roots and expressive.
\subsection{Compounding and reduplication}
There are some compound modifiers in the Tangkhul, which are further reduplicated to denote modified or completely changed meaning. In the process of reduplication, the last syllable of the compound root is partially reduplicated by replacing its initial consonant. The following are some of the most commonly used reduplicated compound adjectives. 
For example: Root + Root + Reduplicate
\begin{itemize}
\item them-reak-sek = them ({\em skill})- reak ({\em pretend})- sek ({\em redu}) \\
Approximate English meaning:  pretending to be very skillful or learned

\item Khon-zar-tar = Khon ({\em sound})- zar({\em dense})- tar ({\em redu}) \\       
Approximate English meaning: very noisy.
                                                              
\end{itemize}

\section{Corpus preparation and preprocessing}
As the existence of the language is very rare in web, we have manually collected/typed few texts 
for our experiment. The current version of the corpus contains 21713 words, out of which 7362 are unique words (include inflections). The articles or segment of articles in the corpus can be categoried into biography (4 numbers), short story (6 numbers), essay (11 numbers), drama (2 numbers) and letter (1 number) with average (approximately) 904 words per article. 
%
%
The first ten frequent words in the corpus are tabulated in Table~\ref{freq10}. 

\begin{table}[!htbp]
\centering
\begin{tabular}{@{}llll@{}} 
\hline
Sr No. & Word & Meaning & frequency\\
1 & eina & 			{\em with}&708 \\
2 &  hi & 			{\em this}&367 \\
3 &  chi & 			{\em that}&358\\
4 &  kala & 		{\em and}&336\\
5 &  da & 			{\em } &184\\
6 &  haowa & 		{\em }&130\\
7 &  kaji & 		{\em that} &122\\
8 &  \~{a}kha & 		{\em one} &122\\
9 &  \~{a}wui & 		{\em his}&120\\
10&  chili & 		{\em there}&110\\
\hline
\end{tabular}
\caption{\label{freq10}Ten most frequent words in the corpus.}
\end{table}

For our experiment, we use Morfessor \cite{creutz+05} version 1.0\footnote{http://www.http://morpho.aalto.fi/projects/morpho; Access date: 30 August, 2017.}, an unsupervised language independent morphology learning package to discover the regularities behind the word formation process. This leaning approach discovers the primitive morphological units such as root/base/stem\footnote{Though, there are differences in the concept of root, base and stem, for this experiment we are using these terms interchangeably to indicate stem}, suffix and prefix of the utterances of a language. As it is based on words or utterance of the language and its frequency, most of the time, the tool discovers prefix (in sequence), stem and suffix (in sequence). During experiment, we found that, a word may have maximum seven different morpheme, mostly sequence of affixes. The morpheme frequency with example is tabulated in 
Table \ref{freqCompo}. We also found that The corpus contains a good number of borrowed words from English and neighbouring languages. 

\begin{table*}[!htbp]
\centering
\begin{tabular}{@{}lll@{}} 
\hline
Words with no affix &  1772& Example: ãkhana\\
Words with one affix &  3600& Example: advocate+la\\
Words with two affixes & 1487  & Example: \~{a}+thingreira+wui\\
Words with three affixes & 410  & Example: kajui+kha+nem\\
Words with four affixes & 78 & Example: ã+lung+th+ung+li\\
Words with five affixes & 12& Example: la+ng+da+ng+l\~{a}+na\\
Words with Six affixes & 3& Example: kha+nga+p+eo+bing+li+la\\
\hline
\end{tabular}
\caption{\label{freqCompo} Affix frequency distribution in the corpus}
\end{table*}

The following examples shows the morphological richness of the considered language segmented using Morfessor.\\ 
maphaning ({\em not thinking}) = ma ({\em no}) + phaning ({\em thinking}) \\
maphaninsali  = ma + pha + nin + sa + li\\
map\~{a}msangmara = ma + p\~{a} + m + sa + ng + mara\\
khamashash\~{a}li = kha + ma + sha + sh + \~{a}li\\
kakhararkhangazai = ka + kharar + kha + nga + zai \\
\~{a}ch\~{a}honthangcham = \~{a} + ch\~{a} + hon + thang + cham\\

\noindent
The following are few examples, the Morfessor segmented correctly.\\ 
a	+	cham	+	\~{a}ram\\
\~{a}	+	lung	+	th	+	ung	+	li\\
\~{a}	+	ng	+	\~{a}	+	ng	+	nao	+	li\\
a	+	ri	+	shang	+	li \\

\noindent
Table \ref{error} tabulated few words incorrectly segmented by Morfessor.  

\begin{table}[!htbp]
\centering
\begin{tabular}{@{}ll@{}} 
\hline
Wrong (By Morfessor) & Correct \\
\~{a} + k\~{a} + khare + wui & \~{a} + k\~{a} + kha + re + wui \\
\~{a} + mathen + pai + ra & \~{a} + ma + then + pai + ra\\
\~{a} + ngas\~{a}m + khuiya & \~{a} + nga + s\~{a}m + khui + ya\\
\hline
\end{tabular}
\caption{\label{error} Incorrect segmentation of words}
\end{table} 

Interestingly, during segmentation, based on the evidence in the corpus, the tool is splitting the following English words used as loan words in Tangkhul (Table~\ref{freqx}). 

\begin{table}[!htbp]
\centering
\begin{tabular}{@{}llp{3cm}@{}} 
\hline  
acqui + red & Wrong\\
activ + ities & Wrong \\
activ + ity & Wrong\\
administra + tion + wui  & Wrong\\
\hline
\end{tabular}
\caption{\label{freqx} Segmentation of loan words}
\end{table}

In the context of loan words, another interesting example we found is {\em acid + p\~{a} + va}. Though the root (the first morpheme) is a valid loan word, the context actually was indicating the root {\em aship\~{a}va} (means {\em wife}).

\section{Conclusion}
The demand for localization through electronic content has made the globe a smaller place and any digital divide in this regard has to be reduced through the inclusion of more naturally occurring languages through the information technology revolution. Towards this direction, the present task is a step to bring Tangkhul into the language technology revolution. In the present work, we reported the morphological processing work of Tangkhul using an unsupervised approach. We found the result of the experiment to be reasonably good as compared to the size of the corpus used in the work. Our future direction includes more experiment on this by including language specific rules and other semi-supervised approaches.

\bibliography{regicon}
\bibliographystyle{acl_natbib}

\end{document}